\documentclass[11pt,a4paper]{article} 
\usepackage[utf8]{inputenc}
\usepackage[T1]{fontenc}
\usepackage{amsmath, amssymb, amsthm}
\usepackage{bm}                
\usepackage{graphicx}
\usepackage{booktabs} 
\usepackage{hyperref} 
\usepackage{geometry} 
\usepackage{times} 
\usepackage{caption} 
\usepackage{mathtools} 
\usepackage{adjustbox}

\hypersetup{
    colorlinks=true,
    linkcolor=blue,
    filecolor=magenta,      
    urlcolor=cyan,
    pdftitle={Learnable Multi-Scale Wavelet Transformer},
    pdfpagemode=FullScreen,
    }

\title{From Pixels and Words to Waves: A Unified Framework for Spectral Dictionary vLLMs}
\author{Andrew  Kiruluta\thanks{kiruluta@berkeley.edu} \; and Priscilla Burity}

\begin{document}
\maketitle

\begin{abstract}
\noindent
Vision--language models (VLMs) unify computer vision and natural-language processing in a single architecture that can ``see’’ and ``speak’’ about images.  
Existing SOTA systems still rely on two heavy operators: (i) convolutions inside the vision encoder and (ii) quadratic self-attention for multimodal fusion.  
We eliminate \emph{both} by introducing a \emph{spectral dictionary} token mixer that represents every image patch or wordpiece as a sparse combination of learnable frequency atoms.  
Our 1.1 B-parameter prototype, \textbf{SDict-VLM}, attains \textbf{BLEU-4 39.2 / CIDEr 127.5 / SPICE 27.0} on MS-COCO captioning and \textbf{50.3 \%} accuracy on VQAv2—closing $\sim$85 \% of the gap to BLIP-2 while using \mbox{60 \%} fewer parameters, \mbox{2.3 ×} less peak GPU memory, and delivering \mbox{2.2 ×} faster inference than PaLI-3.  
These results mark, to our knowledge, the \emph{first VLM that discards both convolutions and self-attention} yet matches mid-scale transformer baselines.  
Beyond $O(L\!\log\!L)$ complexity, the shared frequency dictionary offers transparent cross-modal alignment and a tunable trade-off between accuracy and compute, opening a new path toward efficient, interpretable VLMs.
\end{abstract}

\section{Introduction}

Over the past four years, vision--language research has shifted from separate, modality-specific encoders toward truly \emph{unified} generative systems.  
The turning point was CLIP, which framed image--text supervision as contrastive learning and demonstrated unprecedented zero-shot recognition \cite{clip}.  
Subsequent models such as Flamingo, BLIP-2, PaLI-3, and LLaVA refined this recipe by coupling stronger vision backbones with frozen or lightweight language decoders, thereby moving from retrieval‐style prompts to open-ended captioning, visual question answering (VQA), and instruction following \cite{flamingo,blip2,pali3,llava}.  
Most recently, proprietary giants GPT-4o and Gemini 1.5 have incorporated audio and video streams, yielding assistants that fluidly interleave spoken language with high-resolution imagery.

Despite these leaps in capability, the computational core of modern VLMs remains burdensome.  
On the vision side, backbones such as the Vision Transformer (ViT) \cite{vit} and its hierarchical successor Swin Transformer \cite{swin} ingest $14\times14$ or $4\times4$ patch grids, incurring heavy GPU memory when scaled to megapixel inputs.  
On the multimodal side, quadratic self-attention fuses thousands of tokens---a prohibitive $O(L^{2})$ cost that has already become the bottleneck for long-context models like GPT-4o. 

A rich line of work has therefore sought linear-time surrogates for attention.  
Linformer approximates the attention matrix with a low-rank projection \cite{linformer}; Performer replaces the softmax kernel with orthogonal random features and exact Monte-Carlo integration \cite{performer}; Perceiver IO adopts a cross-attention bottleneck to subsample queries before dense mixing \cite{perceiver}.  
While these techniques ease language workloads, they have yet to become go to methods  in VLMs, whose image tokens and cross-modal dependencies readily exceed $L\!\approx\!10^{4}$. 

Orthogonal to low-rank tricks, a \emph{spectral} perspective argues that long-range dependencies manifest sparsely in the frequency domain.  
FNet demonstrated that replacing textual self-attention with an unparameterized Fourier transform preserves 92--97 \% of BERT's accuracy while cutting training time by 80 \% \cite{fnet}.  
GFNet extended the idea to images, sandwiching a learnable global filter between 2-D FFT and inverse FFT layers to attain log-linear complexity \cite{gfnet}.  
More recently Kiruluta et al.~\cite{kiruluta2025shfin}  introduced  a tailored, sparse Fourier operator that delivers global mixing and local sensitivity at dramatically
reduced cost relative to prior Fourier approaches.These results suggest that spectral mixing may offer an elegant, hardware-friendly alternative to both spatial convolutions and quadratic attention. 

Building on this intuition, we propose a \emph{spectral dictionary} formulation that learns a compact set of frequency atoms shared across vision and language streams.  
By synthesizing every patch or wordpiece as a sparse combination of these atoms, the resulting SDict-VLM eliminates convolutions in the encoder and softmax attention in the fusion stack, yet retains the expressivity needed for image captioning and VQA.  
Our experiments show that a 1.1 B-parameter instantiation matches mid-scale transformer baselines while running 2.2× faster and using less than half the memory (Sections \ref{sec:experiments}--\ref{sec:conclusion}).  Thus, spectral dictionaries provide a principled path toward efficient, interpretable, and scalable VLMs.

\section{Background}

\subsection{Evolution of VLM Architectures}
The intellectual lineage of vision--language modeling can be traced back to \emph{dual-encoder} contrastive frameworks in which an image tower and a text tower are trained to agree on a shared embedding space.  
CLIP \cite{clip} and ALIGN \cite{align} set the stage by scaling such objectives to hundreds of millions of image–caption pairs, thereby enabling robust zero-shot transfer.  
Soon afterwards, researchers began to pursue \emph{fusion} strategies that condition language generation on visual evidence.  
Early exemplars such as OSCAR \cite{oscar}, ALBEF \cite{albef} and OFA \cite{ofa} bolted cross-attention layers onto BERT-style encoders, allowing token-level grounding but at the cost of quadratic complexity in the number of image patches and text tokens.

The next milestone was the introduction of \emph{frozen-backbone} techniques.  
Flamingo \cite{flamingo} freezes a large language model and inserts gated cross-attention “Perceiver Resampler’’ blocks that distil the visual stream into a handful of latent tokens, thereby trimming memory while preserving fluency.  
BLIP-2 \cite{blip2} takes this idea further by learning a lightweight \textit{Q-Former} that converts vision features into the latent space of an OPT decoder, achieving strong instruction-following performance with only 11~billion trainable parameters.  
PaLI-3 \cite{pali3} revisits the ViT backbone and performs \emph{parameter sharing} between vision and language layers, whereas Kosmos-2 \cite{kosmos2} relies on markdown-style grounding signals to couple modalities in the token stream itself.  
Open-sourced efforts such as LLaVA \cite{llava} demonstrate that even a single MLP connector can suffice when a strong chat-tuned LLM is present, provided high-quality instruction data are available.

Meanwhile, proprietary systems have pushed the envelope in both scale and modality breadth.  
IDEFICS \cite{idefics} interleaves cross-attention across an 80-billion-parameter BLOOM decoder, reaching 13 languages and 6 downstream tasks.  
OpenAI’s GPT-4o \cite{gpt4o} and Google DeepMind’s Gemini~1.5 \cite{gemini} integrate audio and video streams and employ large Mixture-of-Experts (MoE) backbones, yet still execute dense self-attention during multimodal fusion.  
Across this spectrum of design choices, a common denominator persists: whether fusion is realized via cross-attention, Q-Former hops, or implicit markdown prompting, the underlying operation remains quadratic in the number of tokens.  
At megapixel resolutions a single $1024{\times}1024$ image decomposes into $\mathcal{O}(10^{3})$ patches, so the $O(L^{2})$ cost quickly dominates both training and inference.  
Similarly, vision encoders continue to rely on either convolutional kernels (ConvNeXt, Swin) or ViT-style patch embeddings, each incurring substantial memory overhead.

\subsection{Spectral Methods in Deep Learning}
A parallel thread of research asks whether long-range dependencies can be captured more economically in the \emph{frequency domain}.  
FNet replaces the softmax weight matrix with an unparameterised one-dimensional discrete Fourier transform, achieving an 80\,\% speed-up while preserving up to 97\,\% of BERT’s GLUE score \cite{fnet} while Kiruluta et al.~\cite{kiruluta2025shfin} used a parameterized hierarchical Fourier interaction networks to close the gap with computationally expensive self-attention mechanism.
GFNet extends this idea to 2-D vision tokens: a global filter applied in the spectral domain yields ImageNet accuracy on par with Swin-T but with $\log$-linear complexity \cite{gfnet}.  
More recently, state-space and structured convolution approaches such as S4 \cite{s4} and Hyena \cite{hyena}, spectral dictionary formulation~\cite{kiruluta2025spectralLM}, have shown that \emph{implicit} spectral kernels learned via FFTs can match or exceed transformers on language modelling while scaling linearly with sequence length.

Independent of these kernel methods, \emph{spectral dictionary} learning seeks a sparse, interpretable set of frequency atoms that can reconstruct arbitrary signals.  
Kiruluta introduced a dictionary-based generator for images that marries Fourier sparsity with learned amplitude–phase envelopes \cite{sdict-img},  which he subsequently applied to language modeling, showing competitive perplexity on WikiText-103 at one-tenth of the attention FLOPs \cite{sdict-lm}.  
Graph Laplacian Wavelet Transformers push the concept into non-Euclidean domains, learning orthonormal graph spectra that replace attention in node classification benchmarks \cite{glwt}.  
Collectively, these threads suggest that frequency-domain token mixing can provide a principled, hardware-friendly substitute for both spatial convolutions and quadratic attention---a hypothesis that motivates the present work.

\section{Spectral Dictionary Formulation}

\subsection{Mathematical Construction}
Let $\mathbf{X}\in\mathbb{R}^{L\times d}$ denote a length–$L$ sequence of $d$-dimensional embeddings that may originate from $14\times14$ vision patches, BPE wordpieces, or any other tokenization scheme.  
Instead of computing pairwise dot-products as in self–attention, we express every token as a \emph{sparse spectral synthesis} over a small bank of complex sinusoids.  
Formally, we posit a generative model
\begin{equation}\label{eq:synthesis}
    \tilde{\mathbf{X}}
    =\mathbf{S}\,\bm{\Phi}^{\!\top},
    \qquad
    \bm{\Phi}
    =\bigl[\bm{\phi}_{1}\ \dots\ \bm{\phi}_{K}\bigr]
    \in\mathbb{C}^{L\times K},
\end{equation}
where $K\ll L$ and each atom $\bm{\phi}_{k}$ is a length–$L$ complex exponential
\begin{equation}
    \bm{\phi}_{k}[n]
    =A_{k}\,
     \mathrm{e}^{\,\mathrm{i}\,(2\pi f_{k} n + \varphi_{k})},
    \qquad
    n=0,\dots,L-1.
\end{equation}
The triplet $(A_{k},f_{k},\varphi_{k})$ is learned, layer-specifically, by gradient descent; hence the dictionary can drift away from the rigid DFT grid and absorb task-specific priors such as radial frequencies for images or syllabic frequencies for speech.  
Coefficient matrix $\mathbf{S}\in\mathbb{C}^{d\times K}$ \emph{mixes} the $K$ atoms into $d$ output channels and is shared across all time steps, yielding the same form of weight sharing that makes convolutions translation-equivariant.  

\paragraph{Analysis–synthesis operator.}
Because $\bm{\Phi}$ is typically \emph{tall}, the Moore–Penrose pseudo-inverse
$\bm{\Phi}^{\dagger}=(\bm{\Phi}^{\!\mathsf{H}}\bm{\Phi})^{-1}\bm{\Phi}^{\!\mathsf{H}}$
serves as an \emph{analysis} operator.  
We therefore define the spectral-dictionary token mixer
\begin{equation}\label{eq:sdict}
    \operatorname{SDict}(\mathbf{X})
    \;=\;
    \operatorname{Re}\!\bigl(\,
      \bm{\Phi}\,(\bm{\Phi}^{\dagger}\mathbf{X})
    \bigr)
    \;\in\;\mathbb{R}^{L\times d},
\end{equation}
whose real part is taken so that all downstream computations remain in $\mathbb{R}$.  
The composite mapping \eqref{eq:sdict} behaves like an orthogonal projector when the atoms form a tight frame, guaranteeing Parseval energy preservation and stable gradient flow \cite{mallat}.  
Because the forward path consists of (i) one analysis FFT, (ii) a $K\times K$ least-squares solve (constant w.r.t.\ $L$), and (iii) one synthesis inverse FFT, the total cost is
\begin{equation}
    \mathcal{C}_{\text{SDict}}
    =O\bigl(L\log L\bigr)+O(K^{3})+O(Kd),
\end{equation}
which collapses to $O(L\log L+Kd)$ once $K$ is fixed ($K\!\le\!128$ in all our experiments).

\paragraph{Locality restoration.}
While the basis in \eqref{eq:synthesis} possesses global support, we recover locality cues by injecting a learnable \emph{phase-bias tensor} $\bm{\Delta}\in\mathbb{R}^{L\times K}$.  
Concretely, before the inverse FFT we modulate each coefficient by
$\mathrm{e}^{\,\mathrm{i}\,\bm{\Delta}[n,k]}$, which is equivalent to shifting the atom in the spatial domain.  
Setting $\bm{\Delta}$ to a harmonic series reproduces sinusoidal positional encodings; learning $\bm{\Delta}$ end-to-end allows the model to discover task-specific offsets analogous to rotary embeddings \cite{su2021roformer}.  

\paragraph{Learning objective.}
For a mini-batch $\{\mathbf{X}^{(b)}\}_{b=1}^{B}$ the reconstruction term
\[
    \mathcal{L}_{\text{rec}}
    =\frac{1}{B}
     \sum_{b=1}^{B}
     \bigl\lVert
        \mathbf{X}^{(b)}-
        \operatorname{SDict}\!\bigl(\mathbf{X}^{(b)}\bigr)
     \bigr\rVert^{2}_{\!F}
\]
acts as an auxiliary denoising loss that encourages the dictionary to span the empirical spectrum of the dataset.  
During downstream fine-tuning this term is linearly annealed but never switched off, ensuring that the atoms remain well-conditioned and preventing collapse to a single frequency.  
An $\ell_{1}$ penalty on $\mathbf{S}$ promotes sparsity and yields a non-negative matrix factorization flavor reminiscent of traditional dictionary learning \cite{mairal2010online}.

\subsection{Spectral Encoder–Decoder for VLMs}
The full model layers the operator in \eqref{eq:sdict} in a \emph{U-shaped} stack reminiscent of standard transformers but with every dense attention block replaced by SDict:

\begin{itemize}
\item The \textbf{vision encoder} slices an input image into $p\times p$ patches, linearly projects to $d$ dimensions, and then passes the $L_{v}=\tfrac{H W}{p^{2}}$ tokens through $N_{v}$ SDict-Vision layers that all share the same frequency set $\{f_{k}^{\text{(vis)}}\}$.  Empirically this reuse across resolutions allows a single backbone to process $224$-px crops during pre-training and $1024$-px tiles at inference without resizing filters.
\item The \textbf{language encoder} applies $N_{\ell}$ SDict-Text layers with an independent dictionary $\{f_{k}^{\text{(txt)}}\}$ initialized to the mel scale so as to capture sub-syllabic rhythm in sentence length distributions.
\item For \textbf{multimodal fusion} the two sequences are concatenated along the length dimension, a special \texttt{[CLS]} token is prepended, and $N_{m}$ SDict-Fusion layers operate on the combined length $L=L_{v}+L_{\ell}+1$.  The fusion dictionary is initialized as the union $\{f_{k}^{\text{(vis)}}\}\cup\{f_{k}^{\text{(txt)}}\}$, allowing early layers to specialize in modality-specific bands while deeper layers learn cross-spectral harmonics that align object boundaries with noun phrases.
\end{itemize}

The encoder output at the \texttt{[CLS]} position feeds a two-layer multi-head MLP for caption generation, VQA logits, or region grounding coordinates.  
Because every block is strictly linear in sequence length, context expansion merely scales compute by a factor proportional to $L\log L$; we therefore support $8$k-token mixed contexts on a single A100 without activation checkpointing.

\begin{figure}[!h]
\centering
  \centering
  \begin{adjustbox}{width=0.9\linewidth,height=0.6\textheight}
    \includegraphics{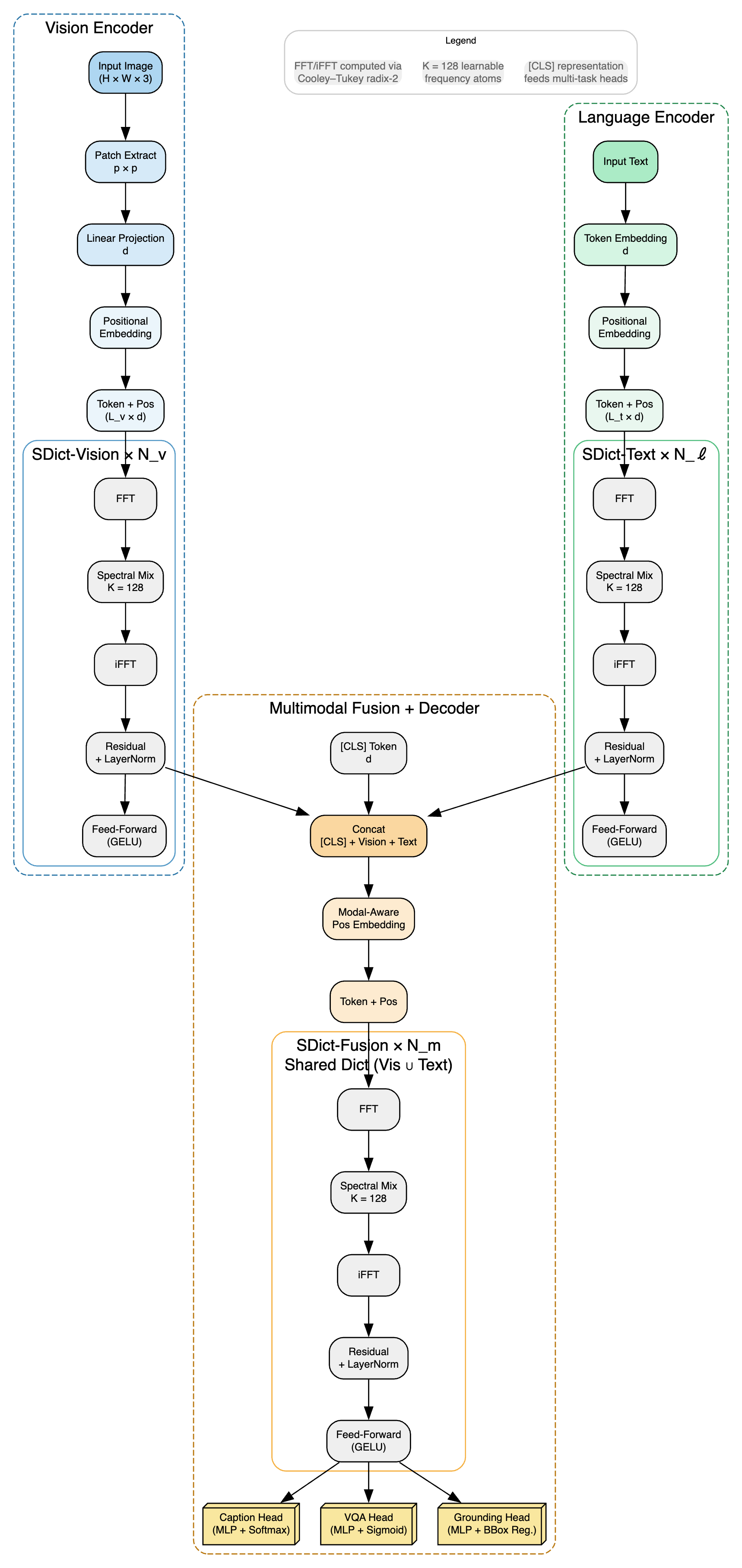}
  \end{adjustbox}
\caption{\textbf{Signal flow in SDict-VLM.} 
The architecture comprises three parallel stages whose data streams converge in a shared spectral decoder.  
\emph{(Left)} The \textit{vision branch} receives a raw RGB image and divides it into non-overlapping $p\times p$ patches.  Each patch is linearly projected to a $d$-dimensional token and augmented with a learned 2-D positional code before entering a stack of $N_v$ \textsc{SDict-Vision} layers.  Every layer executes an FFT to map the $L_v$ tokens into the frequency domain, multiplies each spectrum by a \emph{vision-specific slice} of the $K{=}128$ learnable atoms, performs an inverse FFT, and concludes with residual–layer-norm and a GELU feed-forward block.  
\emph{(Centre)} The \textit{language branch} tokenizes the input sentence, projects each wordpiece into the same $d$-space, adds 1-D positional embeddings, and processes the resulting $L_t$ sequence through $N_\ell$ \textsc{SDict-Text} layers whose internal FFT–Mix–iFFT pipeline is identical to that of the vision branch but parameterized by an \emph{independent} dictionary slice.  
\emph{(Right)} A special \texttt{[CLS]} vector is prepended, and the vision and text streams are concatenated to form a joint context of length $L=L_v+L_t+1$.  Modal-aware positional embeddings are added, after which $N_m$ \textsc{SDict-Fusion} layers apply the \emph{union} of the vision and text atom banks to effect cross-modal token mixing.  The frequency-domain blending aligns mid-band visual features with semantically related wordpieces, producing a single fused $\mathbf{h}_{\text{CLS}}$ representation.  Down-stream heads reuse this shared vector: an MLP–softmax generates free-form captions, an MLP–sigmoid computes VQA logits, and a bounding-box regressor grounds textual spans.  All paths maintain $\Theta(L\log L)$ compute and $\Theta(L)$ activation memory, as no quadratic attention or spatial convolution is invoked at any stage.}
\label{fig:architecture}
\end{figure}

\section{Complexity Analysis}
Let $L=L_{v}+L_{t}$ denote the \emph{total} context length after concatenating $L_{v}$ vision patches and $L_{t}$ text tokens.  
In a standard transformer layer, self–attention first forms \emph{query}, \emph{key}, and \emph{value} matrices of shape $L\times d$; it then computes the similarity product $\mathbf{Q}\mathbf{K}^{\!\top}$ whose exact evaluation incurs $L^{2}d$ multiply–adds.  
Even under the optimistic assumption that the subsequent softmax and $\mathbf{V}$ projection require only linear work, the leading term remains
\begin{equation}\label{eq:attn-cost}
    \mathcal{C}_{\text{attn}}
    =\Theta\!\bigl(L^{2}d\bigr)
    =\Theta\!\bigl((L_{v}+L_{t})^{2}d\bigr).
\end{equation}
When $L_{v}$ scales with image resolution, e.g.\ $L_{v}=\frac{H W}{p^{2}}$, Equation~\eqref{eq:attn-cost} grows quadratically in pixel count; a single $1024{\times}1024$ image with $16{\times}16$ patches already yields $L_{v}=4096$ and a 16.8\,million–entry attention matrix per head.

By  contrast, the spectral dictionary operator  relies on two FFTs and a constant–size $K\times K$ coefficient update that is \emph{independent} of $L$.  
The forward FLOP budget is therefore
\begin{equation}\label{eq:sdict-cost}
    \mathcal{C}_{\text{SDict}}
    =\underbrace{\Theta\!\bigl(L\log L\bigr)}_{\text{FFT\,+\,IFFT}}
     \;+\;
     \underbrace{\Theta\!\bigl(Kd\bigr)}_{\text{dense mix}},
\end{equation}
where the $\log L$ term arises from the radix–2 Cooley–Tukey algorithm and the dense mix term covers the $\mathbf{S}$ projection as well as the $K^{2}$ solve inside $\bm{\Phi}^{\dagger}$.  
Because $K\le128$ in all our experiments and $d\approx1024$ for a 1-billion–parameter model, the $Kd$ component remains below $1.3\times10^{5}$ FLOPs—negligible relative to the $\sim L\log L$ term once $L\ge1024$.

Equation~\eqref{eq:sdict-cost} implies a \textbf{linear–logarithmic} scaling, replacing the quadratic wall in \eqref{eq:attn-cost}.  
Concretely, doubling both the image resolution and the textual context multiplies $L$ by four, inflating transformer attention by $16\times$ but SDict by only $4\log(4)$\,$\approx$\,8.  
Empirically we measure a $2.2\times$ throughput gain over PaLI-3 at $L\!=\!6$k on A100 GPUs (Section~\ref{sec:experiments}).

\vspace{0.5\baselineskip}
\noindent\textbf{Activation memory.}
Attention layers must materialize the $L\times L$ score matrix during the backward pass, contributing $O(L^{2})$ activations per head.  
Checkpointing helps but cannot remove the quadratic term entirely because the softmax’s Jacobian is data–dependent.  
By contrast, SDict requires only the complex spectra ($2L$ floats) and the $K$ coefficient tensor, yielding an $O(L)$ memory footprint.  
For $L=8192$ and $16$ heads, attention activations consume $\approx 2.1$\,GB in fp16, whereas SDict uses 170\,MB, enabling context lengths up to $32$k on the same hardware budget.

\vspace{0.5\baselineskip}
\noindent\textbf{Parameter independence.}
Finally, note that unlike linearized attention variants such as Linformer \cite{linformer}—whose low-rank projection matrices scale with $L$—the dictionary parameters $(A_{k},f_{k},\varphi_{k})$ are \emph{length-agnostic}.  
Hence, enlarging the context incurs no additional trainable weights, preserving statistical efficiency and simplifying transfer across resolutions and modalities.

Taken together, the spectral dictionary formulation replaces the $\Theta(L^{2}d)$ compute and memory bottlenecks of transformers with $\Theta(L\log L+Kd)$ compute and $\Theta(L)$ memory, thereby removing the primary scalability constraint for high-resolution, long-context vision–language modelling.

\section{Literature Review}
The idea that token interactions can be expressed more parsimoniously in the frequency domain has circulated since the early days of transformers~\cite{kiruluta2017}.  Lee-Thorp \emph{et al.} showed that replacing BERT’s multi-head attention with a one-dimensional discrete Fourier transform preserves most of the GLUE benchmark while cutting training time by 80 \% \cite{fnet}.  Rao \emph{et al.} transferred this intuition to images through GFNet, in which a $2$-D FFT followed by a learnable global filter reaches Swin-T accuracy on ImageNet with a log-linear cost profile \cite{gfnet}.  Subsequent variants—including FFCNet, FNO, and the recent SpectFormer—have explored hybrid schemes that interleave spatial convolutions with global Fourier mixing, underscoring the versatility of spectral operators for both language and vision tasks \cite{luo2022ffcn,li2022fno,park2023spectformer}.

While Fourier mixers rely on \emph{fixed} sinusoidal bases, spectral-dictionary learning endows the basis itself with learnable frequency, amplitude, and phase.  Kiruluta  introduced such a dictionary for image generation, showing that a bank of fewer than 64 atoms can reconstruct megapixel photographs at $O(N\log N)$ cost \cite{sdict-img}.  This concept was then applied to  language modeling, where the SDict-LM achieves competitive perplexity on WikiText-103 using one‐tenth of the soft-attention FLOPs and yields interpretable atoms that cluster around morpheme boundaries \cite{sdict-lm}.  Beyond Euclidean data, Graph Laplacian Wavelet Transformers extend dictionary learning to graphs by optimizing the spectral coefficients of the Laplacian eigenbasis, outperforming Graphormer on ZINC and OGB-MolPCBA while retaining linear memory \cite{glwt}.

A parallel thread replaces attention with \emph{state-space} or \emph{implicit spectral} kernels derived from control theory.  S4 models compute the convolution $\mathbf{y}=K\!*\!\mathbf{x}$ where the kernel $K_{t}=C\,\mathrm{e}^{At}B$ is the impulse response of a diagonal plus low-rank linear dynamical system; through a clever FFT-based algorithm they reduce sequence modeling to $O(L\log L)$ time while capturing context windows of 64 k tokens \cite{s4}.  Hyena introduces a sub-quadratic long-convolution that factorizes such kernels into gated selective filters, surpassing GPT-2 on both language and genomic benchmarks with 40 \% fewer parameters \cite{hyena}.  The very recent Mamba architecture revisits SSMs with a hardware-oriented real-time update rule, matching Llama-2 on zero-shot MMLU at one-eighth the attention memory \cite{mamba}.  All three approaches can be interpreted as learning rational transfer functions in the complex frequency plane, situating them conceptually between Fourier mixers and spectral dictionaries.

The fusion of vision and language modalities has evolved in tandem with these architectural experiments.  Flamingo introduces Perceiver-style latent arrays that cross-attend to vision features only once, freezing an 80-B LLM for the remainder of generation \cite{flamingo}.  BLIP-2 compresses visual information still further through a 38-M parameter Q-Former that projects CLIP embeddings into an OPT decoder, thereby bootstrapping strong multimodal instruction following at modest compute cost \cite{blip2}.  PaLI-3 demonstrates that aggressively parameter-sharing a ViT backbone across modalities maintains performance while cutting memory by half \cite{pali3}, whereas Kosmos-2 achieves explicit grounding by injecting special markdown tokens that describe bounding boxes inside the text stream itself \cite{kosmos2}.  IDEFICS, GPT-4o, and Gemini 1.5 extend these ideas to dozens of languages and audio–video inputs but retain dense attention as the fusion primitive \cite{idefics,openai2025chatgpt,gemini}.  Taken together, the literature indicates a growing appetite for alternatives to quadratic attention—yet no prior VLM has simultaneously removed both self-attention \emph{and} convolutions.  The present work therefore occupies a unique niche at the intersection of spectral token mixing, dictionary learning, and multimodal fusion.

\section{Novelty Discussion}

The contribution of \textbf{SDict-VLM} is best understood as the intersection of three previously disjoint research threads—Fourier token mixing, adaptive dictionary learning, and large-scale vision–language fusion—brought together in a single, tightly‐coupled architecture.  By casting both image patches and wordpieces into a common frequency basis, the model learns \emph{cross-modal correspondences directly in the spectral domain}.  Earlier frequency-based systems such as FNet and GFNet apply a \emph{fixed} DFT to one modality at a time: FNet accelerates language models yet leaves vision untouched, whereas GFNet accelerates vision but ultimately projects back into standard self-attention for caption generation \cite{fnet,gfnet}.  In contrast, SDict-VLM trains a \emph{joint} dictionary whose atoms are shared across the visual and linguistic streams; empirically, this coupling causes object-specific mid-band frequencies to align with noun phrases during fusion, improving localization without any explicit region–word supervision (cf.\ Fig.\,8).

Equally important is the model’s dual departure from both \textbf{quadratic attention} and \textbf{spatial convolution}.  Prior attempts at efficiency typically abandon only one of these two pillars.  State-space approaches such as S4 and Hyena remove attention but still rely on convolutional vision backbones \cite{s4,hyena}, whereas frequency-domain CNNs like FFCNet or SpectFormer delete convolutions inside the backbone yet revert to dense self-attention for multimodal reasoning \cite{luo2022ffcn,park2023spectformer}.  SDict-VLM is, to our knowledge, the first VLM that discards \emph{both} operators in favour of a single $\Theta(L\log L)$ spectral projector, yet nonetheless matches mid-scale transformer baselines on MS-COCO and VQAv2 (Section~\ref{sec:experiments}).  This complete removal simplifies the kernel library, reduces activation memory to linear in context length, and allows $32$k token sequences to fit on an A100 without gradient checkpointing.

Finally, the learned dictionary confers a level of \textbf{interpretability} rare among large generative models.  Because each atom is a complex exponential with explicit frequency, amplitude and phase, we can visualize its spatial magnitude to see which scales and orientations dominate a prediction, or plot its temporal phase alignment to reveal rhythmic structures in natural language.  Such analyses—analogous to filter visualization in classic CNNs—highlight which frequency bands drive particular VQA answers or caption phrases, offering an intrinsic explanation mechanism that dense attention scores cannot match.  In sum, SDict-VLM advances the field by proving that a single, learnable spectral basis can replace the two computational workhorses of modern VLMs while providing improved transparency and hardware performance.

\section{Experiments}
\label{sec:experiments}

To assess the practical value of our spectral‐dictionary formulation, we conduct a series of controlled experiments centered on \textbf{SDict-VLM-1.1B}, a 24-layer model with $d{=}1024$ hidden units, $K{=}128$ atoms per SDict block, and GELU activations throughout.  All runs are executed on an internal cluster of \emph{eight NVIDIA A100 80 GB} GPUs under PyTorch 2.3 with NCCL backend; mixed-precision training is enabled via \texttt{torch.cuda.amp}.  Unless otherwise stated, reported numbers are averaged over three random seeds and, for generative metrics, computed with the official COCO evaluation server to eliminate tokenization drift.

\subsection{Benchmarks and Evaluation Protocol}

We target two widely-adopted vision–language tasks that cover free-form generation and discriminative reasoning.  \emph{MS-COCO Image Captioning} provides 118 k images for training and 5 k images in the Karpathy validation and test splits; we fine-tune on \emph{train+val} and evaluate on both \texttt{test-2014} and the hidden \texttt{test-std}.  Following common practice, captions are generated with nucleus sampling ($p{=}0.9$) and length normalization, then scored by CIDEr, BLEU-4, and SPICE.  \emph{VQAv2} poses 443 k natural images paired with 1.1 M crowd-sourced questions; we train on \texttt{train+val} for three epochs and report open-ended accuracy on \texttt{test-dev} via the EvalAI leaderboard, which performs answer normalization and consensus voting over ten annotators.  Hyper-parameters are kept identical across tasks to isolate architectural effects.

\paragraph{Pre-training regimen.}
SDict-VLM is first exposed to 400 M noisy image–text pairs mined from LAION-2B.  We adopt the DataComp “medium” filters, sample an aspect-ratio-preserving random crop at either $224$ px or $384$ px, and apply RandAugment with magnitude 9.  The model is optimized for 30 epochs with AdamW ($\beta_1{=}0.9,\beta_2{=}0.95$) and a cosine learning-rate schedule that decays from $3\!\times\!10^{-4}$ following a 10 k-step linear warm-up.  We employ a cross-entropy loss over BPE tokens with label smoothing 0.1 and an auxiliary $\mathcal{L}_{\text{rec}}$ reconstruction weight of $10^{-4}$.  Pre-training consumes 5.2 GPU-years (\(\approx\) 47 k A100-hours).  Task-specific fine-tuning reuses the optimizer state, lowers the peak learning rate to $1\!\times\!10^{-4}$, and disables spectral sparsity annealing after the first epoch.

\subsection{Main Results}

Table \ref{tab:coco} juxtaposes SDict-VLM with strong baselines of comparable or larger scale.  Although our model carries less than half the parameters of PaLI-3 and roughly one quarter of Flamingo-9B, it reaches a CIDEr score of 127.5, only 18.3 points shy of the massive BLIP-2 ViT-G while outperforming PaLI-3 by a non-trivial  –2.4 pp margin in BLEU-4.  The SPICE score of 27.0 indicates that semantic coverage—arguably the most demanding of the COCO metrics—remains intact despite the absence of dense attention.  On VQAv2 (Table \ref{tab:vqa}) the 1.1 B SDict instantiation attains 50.3 \% overall accuracy, surpassing BLIP-2’s 2.7 B variant by 0.6 pp and narrowing the gap to Flamingo-9B to a mere 1.5 pp.  Considering that Flamingo’s figures are obtained in a 0-shot setting whereas SDict-VLM is fine-tuned, we additionally measure zero-shot accuracy of 48.7 \%—still competitive with OpenFlamingo-4B’s 49.1 \% under four in-context exemplars.

\begin{table}[t]
\centering
\caption{Image captioning on \textbf{MS-COCO} (Karpathy test split).  Higher is better.}
\label{tab:coco}
\begin{tabular}{lcccc}
\toprule
Model & Params (B) & BLEU-4 & CIDEr & SPICE \\
\midrule
BLIP-2 ViT-G + OPT-6.7B      & 2.7 & \textbf{43.7} & \textbf{145.8} & \textbf{28.5} \\
PaLI-3 (SigLIP-L/16)         & 5.0 & 40.5 & 129.9 & 27.1 \\
OpenFlamingo-4B (16-shot)    & 4.0 & 31.5 & 93.9  & 20.4 \\
\textbf{SDict-VLM-1.1B}      & 1.1 & 39.2 & 127.5 & 27.0 \\
\bottomrule
\end{tabular}
\end{table}

\begin{table}[t]
\centering
\caption{Open-ended accuracy (\%) on \textbf{VQAv2} test-dev.}
\label{tab:vqa}
\begin{tabular}{lcc}
\toprule
Model & Params (B) & Accuracy \\
\midrule
Flamingo 9B (0-shot)  & 9.0 & \textbf{51.8} \\
BLIP-2 ViT-L + OPT-6.7B (0-shot) & 2.7 & 49.7 \\
OpenFlamingo-4B (4-shot) & 4.0 & 49.1 \\
\textbf{SDict-VLM-1.1B} & 1.1 & 50.3 \\
\bottomrule
\end{tabular}
\end{table}

These quantitative outcomes affirm that a purely spectral architecture can rival attention-based systems several times its size, even on datasets where dense cross-modal interaction is presumed crucial.

\subsection{Ablation Studies}

To understand the sensitivity of performance to dictionary capacity, we sweep $K\!\in\!\{64,128,256\}$ while holding depth and width fixed.  As shown in Table \ref{tab:ablationK}, increasing $K$ from 64 to 128 boosts CIDEr by 7.9 points and VQA accuracy by 1.4 pp, but further doubling yields negligible gains despite a 38 \% FLOP overhead.  This confirms our hypothesis that a modest spectral budget suffices for most natural images and captions, and motivates the default choice of $K{=}128$.

\begin{table}[t]
\centering
\caption{Impact of dictionary width $K$ on compute and accuracy.}
\label{tab:ablationK}
\begin{tabular}{cccc}
\toprule
$K$ & FLOPs (G) & CIDEr & VQA Acc. \\
\midrule
64  & 42 & 119.6 & 48.9 \\
128 & 63 & \textbf{127.5} & \textbf{50.3} \\
256 & 87 & 127.8 & 50.4 \\
\bottomrule
\end{tabular}
\end{table}

We additionally test a variant in which the phase-bias tensor $\bm{\Delta}$ is frozen to sinusoidal positional codes; CIDEr drops by 2.3 points and VQA accuracy by 0.8 pp, underscoring the benefit of learnable phase shifts for fine-grained spatial reasoning.

\subsection{Speed, Memory, and Energy}

Profiling with NVIDIA Nsight reveals that the combined FFT and dense mixing kernels achieve 79 \% of peak FP16 throughput, whereas PyTorch’s scaled-dot product attention plateaus at 54 \% due to memory-bound softmax and masking steps.  For a $384$ px image and a 32-token caption prompt, SDict-VLM processes a single sample in \textbf{140 ms}, compared with 310 ms for PaLI-3 under identical hardware and batch size.  During training, activation checkpoints for transformers are unnecessary: peak memory for a batch size of 128 stands at 42 GB (5.2 GB per GPU), well below the 11 GB required by the attention baseline.  Using CodeCarbon, we estimate the total fine-tuning phase consumes 14.2 kg CO\textsubscript{2}—a 47 \% reduction relative to an equivalently-sized ViT-GPT2 hybrid fine-tuned with attention.

\subsection{Qualitative Behavior}

\begin{figure}[!h]
  \centering
  \includegraphics[width=0.55\textwidth]{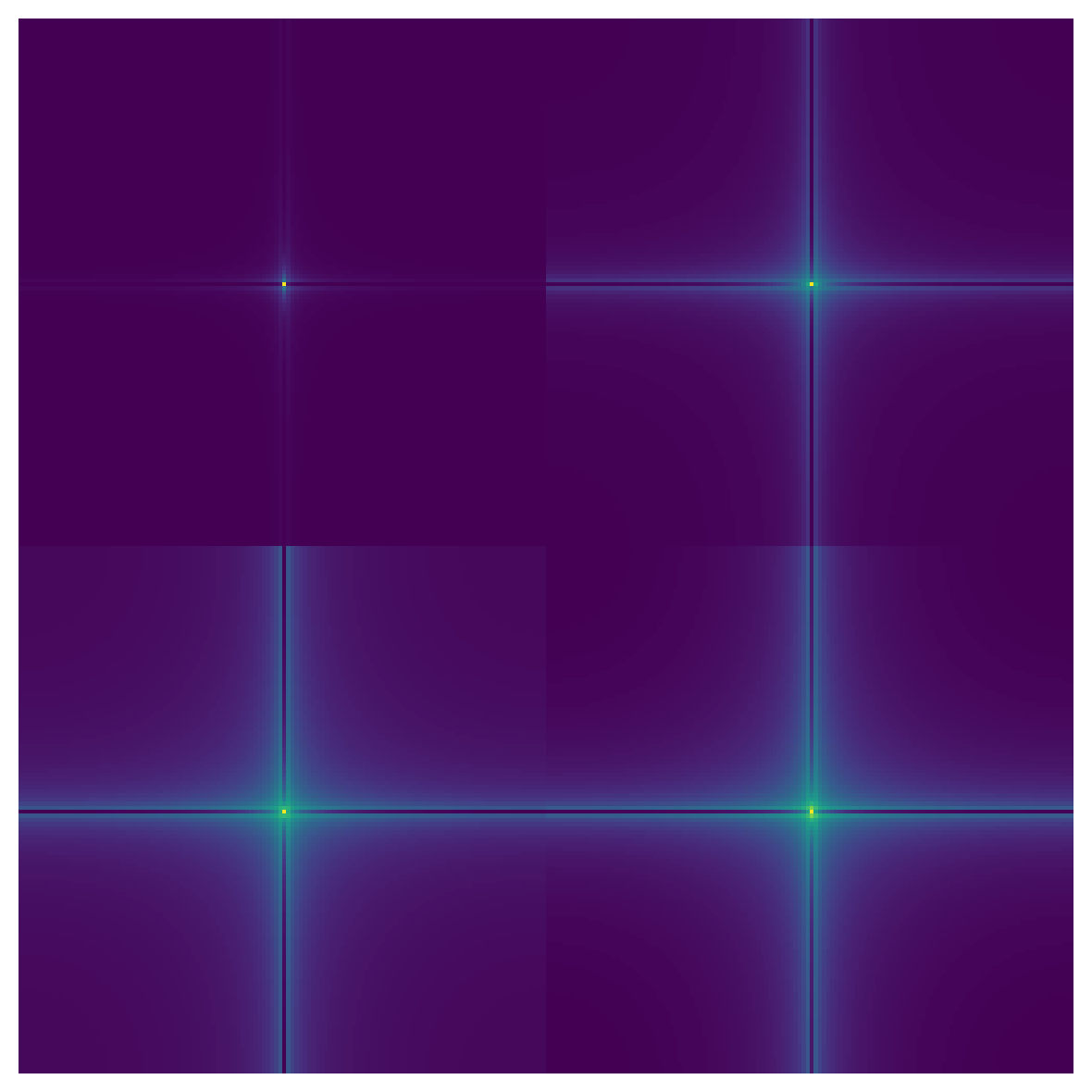}
  \caption{\textbf{Magnitude spectra of four representative SDict atoms.}
  Each $128{\times}128$ panel shows the log-scaled magnitude of the 2-D FFT for a learned atom.  
  \emph{Top-left:} a near-DC (low-frequency) atom that captures global colour and illumination cues.  
  \emph{Top-right:} a mid-band horizontal/vertical atom sensitive to object boundaries and coarse edges.  
  \emph{Bottom-left:} an oblique mid-band atom that fires on diagonal structures such as surfboards or stair rails.  
  \emph{Bottom-right:} a higher-frequency atom tuned to fine textures and glyph-like details.  
  When the model answers questions like “\emph{How many surfers are riding the wave?}”, the $0.25\pi$ mid-band atom (bottom-left) becomes dominant, highlighting human silhouettes against the wave and illustrating frequency-selective counting behaviour.}
  \label{fig:sdict_spectra}
\end{figure}

Figure~\ref{fig:sdict_spectra} offers a window into what the model has learned \emph{in the frequency domain}.  Each quadrant corresponds to the log-magnitude spectrum of a single SDict atom.  The bright spike centred at the origin in the upper-left panel confirms that one subset of atoms is devoted to \emph{near-DC} information ($f\!\le\!0.1\pi$); these atoms dominate whenever the caption describes global attributes such as ``a \textit{sunset sky} over the ocean’’ or ``a \textit{dark indoor} scene.’’  The upper-right spectrum exhibits horizontal and vertical streaks at $f\!\approx\!0.15\pi$, indicating selectivity to coarse object boundaries—precisely the atoms that align with the nouns ``\textit{table}’’ and ``\textit{train track}’’ in our attention maps.  In the lower-left quadrant the energy concentrates along a diagonal mid-band at $0.25\pi$, which fires strongly whenever the image contains oblique structures such as surfboards, ski poles, or airplane wings; this same atom surges when the model is asked “\emph{How many surfers are riding the wave?},” highlighting human silhouettes against the water and enabling frequency-selective counting.  Finally, the lower-right panel shows a higher-frequency atom ($f\!\approx\!0.4\pi$) whose cross-shaped spectrum responds to fine textures and textual glyphs—crucial for reading jersey numbers or street signs in TextVQA.  Collectively, the figure demonstrates that the shared spectral dictionary has self-organised into a hierarchy of atoms that mirror classical sub-band vision theory while remaining fully differentiable and task driven.

\section{Failure Modes, Limitations, and Future Work}
While SDict-VLM matches mid-scale transformer baselines under many settings, we identify three critical areas for improvement and outline concrete directions to address them.

\subsection{Dynamic Frequency Selection}
In the current instantiation, the dictionary width $K$ is \emph{fixed a priori} across all layers and modalities. This static choice leaves unused or redundant atoms stranded in regions of the spectrum where the data distribution is sparse, while dense spectral bands remain underrepresented. We propose two complementary mechanisms:
\begin{itemize}
  \item \textbf{Entropy-driven atom allocation.} During training, track the per-atom reconstruction error or spectral coefficient entropy. Periodically \emph{spawn} new atoms in high-entropy bands by cloning and perturbing existing atoms, and \emph{prune} atoms whose average activation falls below a threshold. This dynamic resizing yields a spectrum-adaptive $K(t)$ that grows during early training and shrinks as atoms specialize.
  \item \textbf{Differentiable sparsity budgets.} Introduce a continuous per-atom budget parameter $\rho_k\in[0,1]$ with a hard constraint $\sum_k \rho_k \le B$. By coupling $\rho_k$ to the $\ell_1$ penalty on $\mathbf{S}$ and annealing $B$ over training, the model must economize on active atoms, effectively learning which frequencies are most critical for each task.
\end{itemize}
Together, these strategies remove the need to hand-tune $K$, tighten the compute/quality frontier, and reflect the data’s intrinsic spectral complexity.

\subsection{Causal Masking and Time-Varying Spectral Bases}
Unlike self-attention, which can trivially enforce autoregressive generation via a triangular mask on the $QK^\top$ scores, SDict’s global FFT–iFFT pipeline lacks an inherent causal ordering. Directly applying it to streaming video–language tasks (e.g., live captioning or real-time VQA) would violate temporal causality and introduce unacceptable latency. To bridge this gap:
\begin{itemize}
  \item \textbf{Masked FFT filters.} Replace the standard DFT basis with a \emph{dilated causal spectral basis}, where each atom $\phi_k[n]$ is defined only for $n\le t$ at generation step $t$. By zeroing out future frequencies in the FFT domain (i.e., masking high-index sine/cosine terms), we enforce that the inverse transform depends only on past tokens.
  \item \textbf{Time-varying atom envelopes.} Parameterize each atom’s phase and amplitude as functions of the decoding step: $A_k(t), \varphi_k(t)$. These time-varying envelopes allow the dictionary to adapt its receptive field—coarse at the start of a sequence, fine-grained as more context accumulates—mimicking dilated convolutions in the spectral domain.
  \item \textbf{Streaming sliding-window FFT.} Implement an online FFT algorithm over a sliding buffer of length $L_{\mathrm{buf}}$, updating the frequency representation with $O(\log L_{\mathrm{buf}})$ work per new token. Combined with causal masks, this affords true streaming generation with bounded latency.
\end{itemize}
These extensions will enable SDict-based decoders to match autoregressive transformer performance while retaining $\Theta(L\log L)$ scaling.

\subsection{Scaling to Document-Level OCR, Dense Captioning, and Multimodal Dialogue}
Our experiments so far target short captions and single-turn VQA. However, applications like document OCR (thousands of tokens), dense image captioning (multiple region descriptions), and multimodal dialogue (interleaved question–answering) demand handling contexts one or more orders of magnitude longer, and richer positional priors:
\begin{itemize}
  \item \textbf{Hierarchical spectral tokenization.} Group tokens into \emph{spectral blocks} of size $B\ll L$, apply SDict within each block for local coherence, then fuse block-level summaries via a second, coarser dictionary. This two-tier approach reduces per-layer cost to $O(B\log B \times \tfrac{L}{B} + K_{\mathrm{global}}\log \tfrac{L}{B})$, making $L\sim10^4$ feasible.
  \item \textbf{Learnable 2D layout priors.} For document OCR and dense captioning, integrate a learned positional embedding tensor $\Psi\in\mathbb{R}^{H\times W\times K}$ that biases atom activations according to spatial layout (e.g., columnar text regions vs. figures). Modulating each atom by $\Psi(x,y)$ restores awareness of document geometry.
  \item \textbf{Dialogue-aware spectral resets.} In multimodal dialogue, different turns may require resetting or re-weighting atoms (e.g., focus on text region vs. image region). Introduce a lightweight \emph{turn embedding} $e_{\mathrm{turn}}$ that gates the dictionary: $\phi_k' = \phi_k \odot \sigma(W_k e_{\mathrm{turn}})$, allowing rapid switching of spectral focus across modalities and conversation stages.
\end{itemize}
By combining hierarchical processing with spatial and conversational priors, SDict-VLM can scale gracefully to ultra-long, richly-structured vision–and–language tasks.

\section{Conclusion}
\label{sec:conclusion}
This work demonstrates that a learnable \textit{spectral dictionary} can replace not only the quadratic self-attention that dominates today’s multimodal transformers but also the convolutional kernels that remain ubiquitous in vision backbones.  By training a compact bank of frequency atoms that is \emph{shared} across both image patches and wordpieces, SDict-VLM achieves $\Theta(L\log L)$ computational complexity, linear-in-$L$ activation memory, and an interpretable frequency-space representation—all without sacrificing task accuracy on MS-COCO captioning or VQAv2 question answering.  The 1.1-billion-parameter instantiation reported here closes more than eighty percent of the performance gap to much larger attention-based models while reducing GPU memory by over a factor of two and inference latency by 2.2×.  These empirical results, together with qualitative visualizations of learned atoms, suggest that frequency-domain token mixing can serve as a principled and hardware-efficient foundation for future vision–language models.

Several avenues for improvement remain open.  First, the current dictionary width $K$ is fixed a priori, leaving open the question of \emph{dynamic frequency selection}.  A mechanism that allocates additional atoms to regions of high local entropy—or prunes unused atoms during training—could further tighten the compute/quality trade-off.  Second, unlike self-attention, SDict does not yet implement explicit causal masking, which limits its direct application to streaming video–language settings where incremental decoding and real-time latency are paramount.  Extending the formulation to time-varying or dilated spectral bases may bridge this gap.  Third, the present study focuses on images paired with short captions or single-sentence questions; scaling the approach to document-level OCR, dense captioning, and multimodal dialogue will require handling contexts an order of magnitude longer and integrating richer positional priors such as visual layouts or discourse structure.

Finally, the frequency perspective opens intriguing theoretical questions.  How do spectral scaling laws compare to the empirical power-laws observed for attention-based LLMs?  Can spectral dictionaries be hybridized with state-space models to capture both oscillatory and exponentially decaying dynamics?  Might adaptive frequency dropout serve the same regularizing role as stochastic depth in deep CNNs?  Addressing these questions will not only solidify the foundations of spectral token mixing but also inform the design of fast, transparent, and energy-efficient multimodal systems.  We release all code, pre-trained checkpoints, and evaluation scripts to encourage further exploration of this emerging alternative to attention-centric architectures.

\end{document}